\documentclass[10pt,twocolumn,letterpaper]{article}

\usepackage{wacv}
\usepackage{times}
\usepackage{epsfig}
\usepackage{graphicx}
\usepackage{amsmath}
\usepackage{amssymb}
\usepackage{booktabs}
\usepackage{pifont}
\usepackage{stfloats}
\usepackage{multirow}
\usepackage{booktabs}
\usepackage{caption}
\usepackage{subcaption}
\usepackage{enumitem}

\usepackage[accsupp]{axessibility}  

\def\wacvPaperID{703} 

\wacvalgorithmstrack   

\wacvfinalcopy 

\ifwacvfinal
\usepackage[breaklinks=true,bookmarks=false]{hyperref}
\else
\usepackage[pagebackref=true,breaklinks=true,colorlinks,bookmarks=false]{hyperref}
\fi

\begin{document}

\title{Li3DeTr: A LiDAR based 3D Detection Transformer}

\author{Gopi Krishna Erabati and Helder Araujo\\
Institute of Systems and Robotics\\
University of Coimbra, Portugal\\
{\tt\small \{gopi.erabati, helder\}@isr.uc.pt}
}

\maketitle
\thispagestyle{empty}

\begin{abstract}
   Inspired by recent advances in vision transformers for object detection, we propose Li3DeTr, an end-to-end \textbf{Li}DAR based \textbf{3}D \textbf{De}tection \textbf{Tr}ansformer for autonomous driving, that inputs LiDAR point clouds and regresses 3D bounding boxes. The LiDAR local and global features are encoded using sparse convolution and multi-scale deformable attention respectively. In the decoder head, firstly, in the novel Li3DeTr cross-attention block, we link the LiDAR global features to 3D predictions leveraging the sparse set of object queries learnt from the data. Secondly, the object query interactions are formulated using multi-head self-attention. Finally, the decoder layer is repeated $L_{dec}$ number of times to refine the object queries. Inspired by DETR, we employ set-to-set loss to train the Li3DeTr network. Without bells and whistles, the Li3DeTr network achieves 61.3\% mAP and 67.6\% NDS surpassing the state-of-the-art methods with non-maximum suppression (NMS) on the nuScenes dataset and it also achieves competitive performance on the KITTI dataset. We also employ knowledge distillation (KD) using a teacher and student model that slightly improves the performance of our network.
\end{abstract}

\section{Introduction}
With the advent of deep learning networks for computer vision \cite{resnet,vgg} and large-scale datasets \cite{deng2009imagenet} the research on perception systems for scene understanding of autonomous vehicles is growing rapidly. 3D object detection is one of the key processes of autonomous driving, which is a two fold process of classification and localization of the objects in the scene. LiDAR is one of the significant sensors of autonomous vehicles which provides precise 3D information of the scene. Although there is a huge progress in 2D object detection approaches \cite{detr, duan2019centernet, liu2016ssd,   redmon2018yolov3, fasterrcnn, tian2019fcos}, the CNN-based approaches are not well directly adapted to LiDAR point clouds due to their sparse, unordered and irregular nature. 

Earlier approaches for 3D object detection on LiDAR data can be divided into two types: point-based and grid-based methods. Point-based methods \cite{pointformer, shi2019pointrcnn, yang20203dssd} are based on point operations \cite{qi2017pointnet, qi2017pointnet++} which detect the 3D objects directly from the point clouds. 
Grid-based methods either voxelize the points into volumetric grids or project the points to Birds Eye View (BEV) space. 
The advantage of BEV projection is that it preserves euclidean distance, avoids overlapping of objects and the object size is invariant to distance from ego vehicle which is significant for autonomous driving scenarios. The sparse CNN-based voxel feature extraction \cite{second} is advantageous but it can not extract rich semantic information with limited receptive fields. We mitigate this issue by employing a multi-scale deformable attention \cite{zhu2020deformable} encoder to capture global LiDAR feature maps.

Earlier approaches either use two-stage detection pipeline \cite{fastpointrcnn, shi2019pointrcnn} or anchors \cite{pointpillars, zhou2018voxelnet} or anchor-free networks \cite{wang2021fcos3d, pillarod, centerpoint} for 3D object detection, but all of them employ post-processing method like NMS to remove redundant boxes. Inspired by Object-DGCNN \cite{objectdgcnn}, we formulate the 3D object detection problem as a direct set prediction problem to avoid NMS.

We propose an end-to-end, single-stage LiDAR based 3D Detection Transformer (Li3DeTr) network to predict the 3D bounding boxes for autonomous driving. Firstly, the voxel features are extracted with SECOND \cite{second} by leveraging sparse convolutions \cite{sparseconv} and BEV transformation or with PointPillars \cite{pointpillars}. Secondly, we employ an encoder module with multi-scale deformable attention \cite{zhu2020deformable} to capture rich semantic features and long range dependencies in BEV feature maps to generate LiDAR global features. The LiDAR global features are passed to the decoder module. Finally, we introduce a novel Li3DeTr cross-attention block in the decoder to link the LiDAR global features to the 3D object predictions leveraging the learnt object queries. The object queries interact with each other in multi-head self-attention block \cite{vaswani2017attention}. The object queries are iteratively refined and 3D bounding box parameters are regressed in every decoder layer. Inspired by DETR \cite{detr}, we use set-to-set loss to optimize our network during training.

We conduct experiments on two publicly available autonomous driving benchmarks, nuScenes \cite{caesar2020nuscenes} and KITTI \cite{kitti} dataset. Our network achieves 61.3\% mAP and 67.6\% NDS on the nuScenes dataset surpassing the state-of-the-art CenterPoint \cite{centerpoint} and Object-DGCNN \cite{objectdgcnn} by 3.3\% mAP (and 2.1\% NDS) and 2.6\% mAP (and 1.6\% NDS) respectively.

\noindent Our main contributions are as follows:
\begin{itemize}[noitemsep,topsep=0pt]
    \item We propose an end-to-end, single-stage LiDAR based 3D Detection Transformer (Li3DeTr) for autonomous driving. Our method achieves 61.3\% mAP and 67.6\% NDS on the nuScenes \cite{caesar2020nuscenes} dataset which surpassed state-of-the-art LiDAR based object detection approaches. Our method achieves competitive performance (without NMS) to other approaches (with NMS) on the KITTI \cite{kitti} dataset. Similar to DETR \cite{detr}, our approach does not require NMS, hence it is effective to apply knowledge distillation with teacher and student model to improve the accuracy.
    \item We introduce a novel Li3DeTr cross-attention block to link the LiDAR global encoded features to 3D object predictions leveraging the learnt object queries. The attention mechanism in encoder and decoder helps to detect large-size objects effectively as shown in Table~\ref{tab:apbycat}. The ablation study shown in Table~\ref{tab:attn} justifies our novel Li3DeTr cross-attention block.
    \item We release our code and models to facilitate further research.
\end{itemize}

\section{Related Work}
\label{sec:relatedwork}
The LiDAR point cloud based 3D object detection approaches can be divided into two categories: point-based and grid-based, depending on the type of data representation used to predict the 3D bounding boxes.

\noindent\textbf{Point-based methods} \cite{qi2018frustum, pvrcnn, shi2019pointrcnn, yang20203dssd, yang2019std} directly use the sparse and unordered set of points to predict 3D bounding boxes. The point features are aggregated by multi-scale/multi-resolution grouping and set abstraction \cite{qi2017pointnet, qi2017pointnet++}. PointRCNN \cite{shi2019pointrcnn} employs a two-stage pipeline for 3D object prediction. PVRCNN \cite{pvrcnn} models point-voxel based set abstraction layer to leverage the advantage of point and voxel based methods. Frustum-PointNet \cite{qi2018frustum} uses 2D object detection to sample a frustum of points to apply PointNet \cite{qi2017pointnet} to predict 3D objects. Although point-based methods achieve large receptive fields with set abstraction layer, they are computationally expensive.

\noindent\textbf{Grid-based methods.} As the LiDAR point clouds are sparse and unordered set of points, many methods project the points to regular grids such as voxels \cite{second, zhou2018voxelnet}, BEV pillars \cite{pointpillars} or range projection \cite{tothepoint, fan2021rangedet, sun2021rsn}. The point clouds are discretized into 3D voxels \cite{dss, zhou2018voxelnet} and 3D CNNs are employed to extract voxel-wise features. However, 3D CNNs are computationally expensive and requires large memory, in order to mitigate this problem \cite{hotspotnet, second} use sparse 3D CNNs \cite{sparseconv} for efficient voxel processing. The LiDAR point cloud is projected into BEV map in PointPillars \cite{pointpillars} and PIXOR \cite{yang2018pixor} and 2D CNNs are employed to reduce the computational cost, however such projection induces 3D information loss. In order to mitigate this issue, some methods \cite{second, centerpoint} compute voxel features using sprase convolutions and then project the voxel features into BEV space, and finally predict the 3D bounding boxes in the BEV space. As this approach takes the advantage of voxel and BEV space, we test our network with SECOND \cite{second} and PointPillars \cite{pointpillars} feature extraction networks. In order to achieve large receptive fields similar to point-based methods \cite{qi2017pointnet, qi2017pointnet++},  we model long-range interactions of local LiDAR features using multi-scale deformable attention \cite{zhu2020deformable} block in our encoder to obtain LiDAR global features.

\noindent\textbf{Transformer-based methods.}
Earlier approaches \cite{fastpointrcnn, pointpillars, pvrcnn, shi2019pointrcnn, yang2019std, zhou2018voxelnet} object detection head employ anchor boxes to predict the objects, however anchor boxes involve hand-crafted parameter tuning and they are statistically obtained from the dataset. To mitigate this issue, some approaches \cite{hotspotnet, pillarod, yang2018pixor, centerpoint} followed anchor-free pipeline by computing per-pixel or per-pillar prediction. But these approaches use NMS to remove redundant boxes. DETR \cite{detr} is the first transformer architecture which formulated 2D detection problem as a direct set prediction to remove NMS. Our network follows similar formulation for 3D object detection. Some approaches \cite{votr, pointformer, cwt} used transformer for feature extraction networks. 3DETR \cite{3detr} is a fully transformer based architecture for 3D object detection using vanilla transformer \cite{vaswani2017attention} block with minimal modifications. 3DETR directly operate and attend on points whereas our approach voxelize the points and attend the BEV global voxel features which is computationally efficient for autonomous driving scenarios. 3DETR employs downsampling and set-aggregation operation \cite{qi2017pointnet++} on the input points of indoor scenarios because the computational complexity of self-attention increases quadratically ($\mathcal{O}(n^2)$) with the number of input points. Moreover, 3DETR is effective on indoor datasets, where the points are dense and concentrated. Object-DGCNN \cite{objectdgcnn} employs a graph-based model for transformer-based 3D object detection for outdoor environments. BoxeR \cite{nguyen2022boxer} introduces a novel and simple Box-Attention which enables spatial interaction between grid features. BoxeR-2D enables end-to-end 2D object detection and segmentation tasks, which can be extended to BoxeR-3D for end-to-end 3D object detection. VISTA \cite{deng2022vista} is a plug and play module to adaptively fuse multi-view features in a global spatial context, incorporated with \cite{hotspotnet, centerpoint}.  It introduces dual cross-view spatial attention to leverage the information in BEV and Range View (RV) features. We formulate our model with voxel-BEV based CNN backbone architecture for local feature extraction and an attention-based architecture for global feature extraction to increase the receptive field size and finally a transformer decoder head to link global features and 3D predictions.

\section{Methodology}
\begin{figure*}[htp]
    \centering
    \includegraphics[scale=0.90]{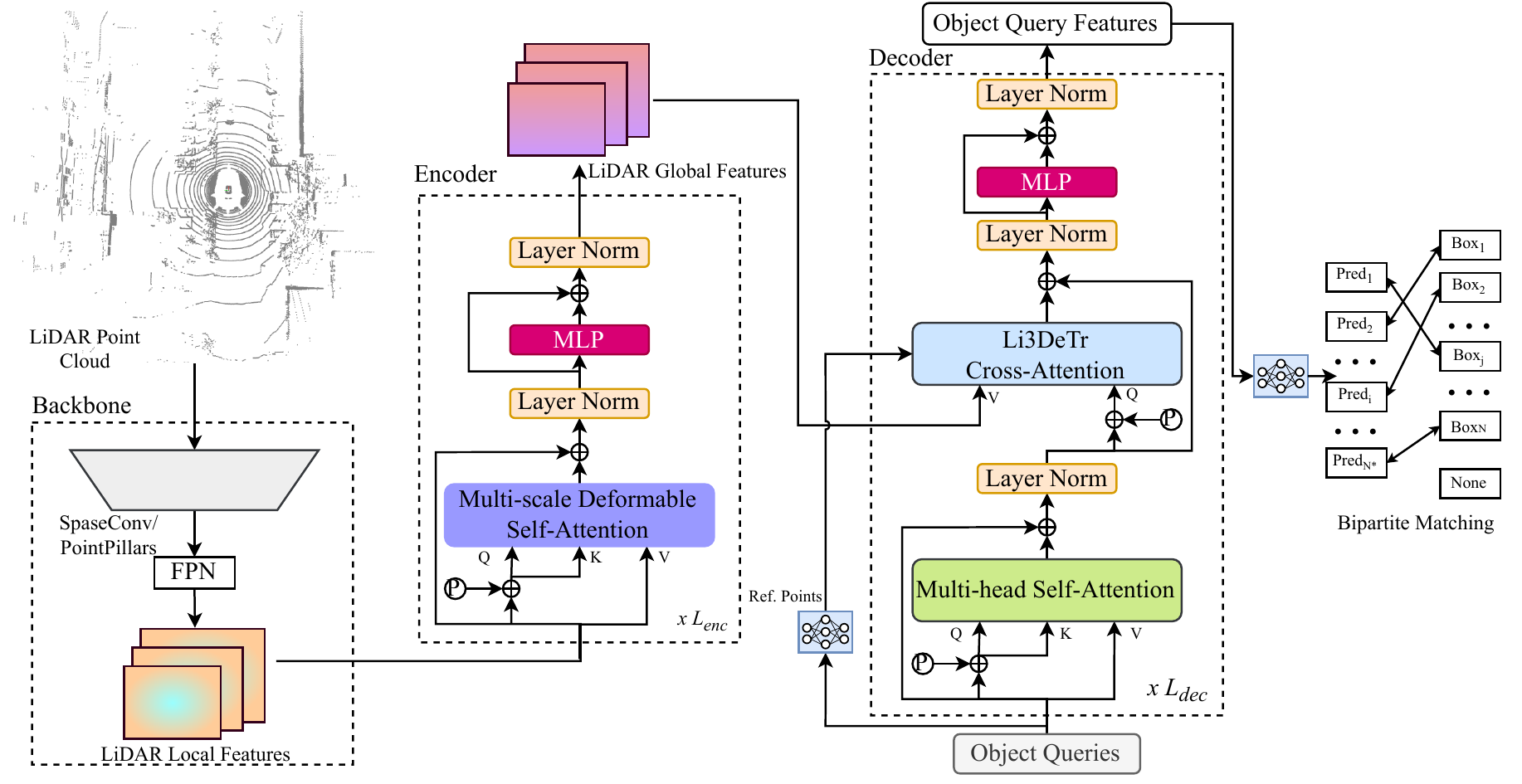}
    \caption{An overview of Li3DeTr architecture. It is an end-to-end, single-stage network which inputs LiDAR point cloud and predicts 3D bounding boxes. The local and global point features are linked to object predictions in transformer encoder-decoder architecture leveraging object queries.}
    \label{fig:li3detr}
\end{figure*}

Our LiDAR based 3D Detection Transformer (Li3DeTr) architecture inputs LiDAR point cloud and predicts 3D bounding boxes in large-scale outdoor environments such as autonomous driving. The network contains two main modules: backbone and transformer encoder-decoder as shown in Figure~\ref{fig:li3detr}. Inspired by the state-of-the-art 3D object detection approaches \cite{sparseconv, pointpillars}, our CNN-based backbone (\S~\ref{subsec:backbone}) module learns grid-based local voxel features. Specifically, we employ BEV grid, not only because the 2D grid-like features are a good trade-off between accuracy and efficiency but also very relevant for autonomous driving, as there is possibly one object on every grid cell on the ground plane. The multi-scale deformable attention \cite{zhu2020deformable} based encoder (\S~\ref{subsec:encoder}) module learns the multi-scale global voxel features. The encoder module alternates between multi-scale deformable attention block and multi-layer perceptron (MLP) block and is repeated $L_{enc}$ number of times. The novel Li3DeTr cross-attention block in the decoder (\S~\ref{subsec:decoder}) module links the global voxel features to the 3D object predictions leveraging the learnt object queries. The object queries interact with each other in multi-head self-attention \cite{vaswani2017attention} block. The decoder module is repeated $L_{dec}$ number of times with alternating multi-head self-attention, Li3DeTr cross-attention and MLP blocks. The refined object queries are transformed into 3D bounding box parameters and the network is trained end-to-end using permutation-invariant loss \cite{detr} (\S~\ref{subsec:loss}).

\subsection{Backbone: Local features}
\label{subsec:backbone}
Our network inputs LiDAR point cloud $\mathcal{P}= \{p_1,\ldots,p_i, \ldots, p_N\}\subset \mathbb{R}^3$. To accelerate the 3D object detection for large-scale point clouds, we scatter the points into BEV grid and use CNNs to extract local point features. We test two pipeline: 1) We voxelize the point cloud with [0.1, 0.1, 0.2] metres voxel size and employ SparseConv \cite{sparseconv} to compute 3D sparse convolutions and obtain local voxel features. The empty voxels are filled with zeros and the sparse voxels are transformed to a BEV 2D grid-like features. 2) We convert the point cloud into a dense BEV pillar map as in PointPillars \cite{pointpillars} with [0.2, 0.2, 8] metres pillar resolution. We use pillar feature net to transform the pillar features. Finally, we employ SECOND \cite{second} backbone to extract local voxel features from the sparse voxel or BEV pillar features and further transform them using a feature pyramid network (FPN) \cite{fpn} to obtain multi-scale local voxel feature maps {$ \mathcal{F}_1, \mathcal{F}_2, \mathcal{F}_3, \mathcal{F}_4 $} where $\mathcal{F}_j \subset \mathbb{R}^{H^j\times W^j\times C^j}$.

\subsection{Encoder: Global features}
\label{subsec:encoder}
In order to obtain global voxel features from the local voxel feature maps, we employ multi-scale deformable attention \cite{zhu2020deformable} mechanism because the conventional attention mechanism \cite{vaswani2017attention} leads to unacceptable computational complexities in encoding high resolution feature maps. The multi-scale deformable attention combines the best of the sparse sampling of deformable convolution \cite{dai2017deformable} and long-range relation framework of transformers \cite{vaswani2017attention}. The input and output to encoder module are multi-scale feature maps with the same resolution. The deformable attention attends only a small set of key sampling points around a reference point and thereby reduces the computational complexity. The reference point in the deformable self-attention for each query pixel is itself. Each encoder layer consist of multi-scale deformable self-attention and MLP blocks with residual connections and is repeated $L_{enc}$ times. The global voxel features extracted from the encoder is passed to Li3DeTr cross-attention block in the decoder.

\subsection{Decoder}
\label{subsec:decoder}
The state-of-the-art 3D object detection approaches either formulates the detection head with dense set of anchor boxes or dense per-pillar prediction such as \cite{pillarod, centerpoint}, followed by NMS. We remove the need of post-processing step (like NMS) by formulating the detection head to predict a set of bounding boxes instead of per-pillar prediction. This is formulated in the decoder as detailed below.

The decoder inputs a set of object queries $\mathcal{Q}^l = \{q_i^l\}_{i=1}^{N_q} \in \mathbb{R}^d$ (where $l \in \{1, 2, \ldots, L_{dec}\}$, $N_q$ is number of queries and $\mathcal{Q}^1$ are learnt with the model weights) and global voxel feature maps $\{\mathcal{F}_j\}_{j=1}^4$ and it consists of decoder layer that is repeated $L_{dec}$ number of times to refine the object queries. 

For the first decoder layer, the 3D reference points are encoded from the object queries using a single-layer fully connected (FC) network and sigmoid normalization as in Equation~\ref{eq:refpoints}. 
\begin{equation}
    r_i = \phi_{ref}(q_i),
    \label{eq:refpoints}
\end{equation}
where $r_i \in [0, 1]^3$ and $\phi_{ref}$ is a FC layer. Each decoder layer consists of Li3DeTr cross-attention block, multi-head self attention block and MLP block with skip connections as shown in Figure~\ref{fig:li3detr}.

\begin{figure*}[htp]
    \centering
    \includegraphics{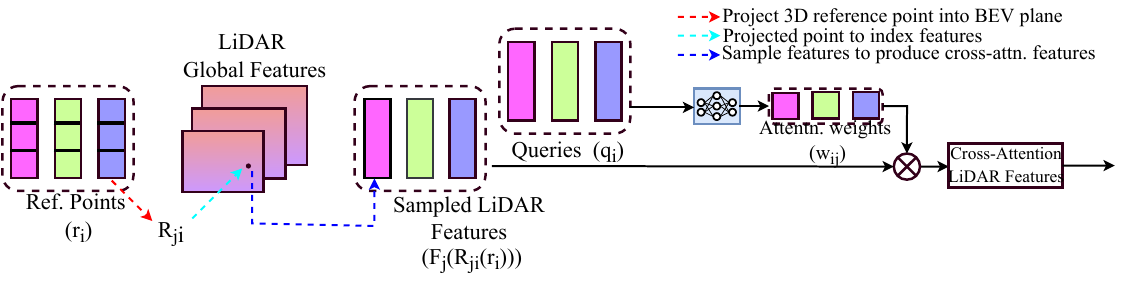}
    \caption{An overview of Li3DeTr cross-attention block}
    \label{fig:li3detrcross}
\end{figure*}

\noindent\textbf{Li3DeTr cross-attention} block inputs the object queries $\mathcal{Q} = \{q_i\}_{i=1}^{N_q}$ (we drop the layer index for simplicity), 3D reference points $r_i$ and LiDAR global multi-scale feature maps $\{\mathcal{F}_j\}_{j=1}^4$. The formulation of Li3DeTr cross-attention block is illustrated in Figure~\ref{fig:li3detrcross}.

Let $\mathcal{R}_{ji}$ represents the transformation for the projection of reference point $r_i$ into scale $j$ of LiDAR global voxel feature map. The LiDAR BEV feature map at scale $j$ is bilinearly sampled at the location of projection of reference point ($\mathcal{R}_{ji}(r_i)$), given by $\mathcal{F}_j(\mathcal{R}_{ji}(r_i))$. The attention weights ($w_{ij}$) for each query $q_i$ at each sampled feature scale $j$ is computed by a FC layer ($\phi_{samp}$) and sigmoid normalization, where $w_{ij} = \phi_{samp}(q_i)$. The sampled features from multi-scale feature maps are added together to obtain cross-attention features ($\mathcal{F}_i^{CAttn}$) for $i$-th reference point as:
\begin{equation}
    \mathcal{F}_i^{CAttn} = \sum_{j=1}^4 \mathcal{F}_j(\mathcal{R}_{ji}(r_i)) . w_{ij}
\end{equation}

\noindent Finally, we update the queries as,
\begin{equation}
    q_i = q_i + \mathcal{F}_i^{CAttn} + PE(c_i),
\end{equation}

\noindent where PE is positional encoding of reference point to make the queries location aware. The queries interact with each other in the multi-head self-attention block and we follow skip connection following \cite{vaswani2017attention}. The object queries ($\mathcal{Q}^l$) are updated in each decoder layer.

 We employ two FC layers $\phi_l^{reg}$ and $\phi_l^{cls}$ to predict offset to box position $\Delta p_i^l \in \mathbb{R}^3$, box size ($l_i^l, w_i^l, h_i^l$), box orientation ($\sin{\theta_i^l}, \cos{\theta_i^l}$), box velocity ($v_{xi}^l, v_{yi}^l$) and class label ($\hat{y}_i^l$) respectively for each object query ($q_i^l$).
 
 We refine the reference points for each query in each decoder layer except for the first decoder layer (which are encoded using Equation.~\ref{eq:refpoints})  by using the predictions of box position in the previous layer, as
 \begin{equation}
     r_i^{l+1} = r_i^l + \Delta p_i^l
 \end{equation}

\subsection{Loss}
\label{subsec:loss}
Similar to \cite{detr, detr3d, objectdgcnn}, the error between predictions and ground-truths is calculated by set-to-set loss. Firstly, the one-to-one assignment between predictions and ground-truths is performed by Hungarian algorithm \cite{kuhn1955hungarian}. Secondly, we use bipartite matching to match predictions and ground-truths. Finally, we employ $L_1$ loss and focal loss \cite{lin2017focal} to calculate bounding box regression and classification loss respectively, given the bipartite matching.

\section{Experiments}
We evaluate Li3DeTr network on the publicly available autonomous driving datasets, nuScenes \cite{caesar2020nuscenes} and KITTI \cite{kitti}. We introduce the experimental setup (\S~\ref{subsec:expsetup}) with dataset details and evaluation metrics (model settings and training details are provided in supplementary), and then present both qualitative and quantitative results (\S~\ref{subsec:results}) and analysis on the nuScenes and KITTI dataset. We perform ablation studies (\S~\ref{subsec:ablation}) to study the different configurations of our network.

\subsection{Experimental Setup}
\label{subsec:expsetup}
\noindent\textbf{nuScenes dataset} \cite{caesar2020nuscenes} contains 750, 150 and 150 sequences (of $\sim$20s duration) with 28K, 6K and 6K annotated samples for training, validation and testing respectively. Each sample consists of 32-beam LiDAR point cloud with 30K points. The dataset also provides 9 non-key frames (called sweeps) to aggregate to one key-frame, resulting in $\sim$300K points per annotated frame. 10 different categories of objects are available to compute the metrics.

\noindent\textbf{Kitti dataset} \cite{kitti} consists of 7481 and 7518 training and testing samples. The training samples are further divided into 3712 \textit{train} and 3769 \textit{val} splits. Each sample consists of 32-beam LiDAR point cloud. Cars, pedestrians and cyclists are the three main categories for evaluation.

\noindent\textbf{Evaluation metrics.}
The two main metrics following the official evaluation of nuScenes dataset \cite{caesar2020nuscenes} are mean average precsion (mAP) and nuscenes detection score (NDS). In addition, we also evaluate true positive (TP) metrics: average translation error (ATE), average scale error (ASE), average orientation error (AOE), average velocity error (AVE), average attribute error (AAE). We follow the official evaluation metric of KITTI dataset \cite{kitti} mean average precision (mAP) with IoU threshold of 0.7 for \textit{car} category and 0.5 for \textit{pedestrian} and \textit{cyclist} categories..

\subsection{Results}
\label{subsec:results}
\subsubsection{Quantitative Results} 
\label{subsubsec:quantres}
We compare Li3DeTr network with the state-of-the-art methods on the nuScenes \cite{caesar2020nuscenes} \textit{test} dataset as shown in Table~\ref{tab:quantres}. Our network surpassed the state-of-the-art CNN-based CenterPoint \cite{centerpoint} by 3.3\% mAP and 2.1\% NDS and transformer-based Object-DGCNN \cite{objectdgcnn} network by 2.6\% mAP and 1.6\% NDS. Albeit CenterPoint \cite{centerpoint} uses post-processing method like NMS, our approach of formulating object detection as a direct set prediction problem inspired by DETR \cite{detr} doesn't require NMS to obtain the gain in mAP and NDS. We also compare with the NMS-free state-of-the-art transformer-based network Object-DGCNN \cite{objectdgcnn} with two different backbones: PointPillars \cite{pointpillars} and VoxelNet \cite{zhou2018voxelnet}. Our transformer based NMS-free approach surpassed in both \textit{pillar} \cite{pointpillars} and \textit{voxel} \cite{zhou2018voxelnet} backbones by 0.6\% mAP (and 0.2\% NDS) and 2.6\% mAP (and 1.6\% NDS) respectively. Although our Li3DeTr network outperforms most of the other methods in terms of mAP and NDS, VISTA-OHS \cite{deng2022vista} performs slightly better than our approach on the nuScenes \textit{test} dataset. VISTA is a plug and play module to fuse multi-view features incorporated with \cite{hotspotnet} which requires post-processing methods like NMS, whereas our approach is a \textit{standalone} method for 3D object detection without NMS. The perfromance of our approach compared with state-of-the-art approaches on nuScenes \textit{val} dataset is provided in supplementary.

We employ knowledge distillation (KD) with a teacher and student model. The earlier works on 3D object detection involve NMS, so it is not effective to distill those models. As our approach is NMS-free, we can effectively distill the information between models with similar detection heads. We train a teacher model with the loss given in \S~\ref{subsec:loss} and then we train a student model (with same architecture as teacher model) with the supervision of output of teacher model and the ground-truth. With the KD, we achieve 62.2\% mAP and 68.0\% NDS which is 0.9\% mAP and 0.4\% NDS improvement over our model without KD.

\begin{table*}[htp]
\centering
\caption{Comparison of recent works on nuScenes \cite{caesar2020nuscenes} \textit{test} set.}
\label{tab:quantres}
\begin{tabular}{@{}cccccccccc@{}}
\toprule
Method  & NDS $\uparrow$ & mAP $\uparrow$ & mATE $\downarrow$ & mASE $\downarrow$ & mAOE $\downarrow$ & mAVE $\downarrow$ & mAAE $\downarrow$ & NMS \\ \midrule 
PointPillars \cite{pointpillars} & 55.0 & 40.1 & 39.2 & 26.9 & 47.6 & 27.0 & 10.2 & \checkmark \\
SSN \cite{zhu2020ssn} & 61.7 & 51.0 & 33.9 & 24.5 & 42.9 & 26.6 & 8.7 & \checkmark \\
CyliNet RG \cite{cylinetrg} & 66.1 & 57.6 & 28.3 & 25.3 & 29.1 & 26.8 & 18.0 & \checkmark \\
CVCNet-ens \cite{cvcnet} & 66.6 & 58.2 & 28.4 & 24.1 & 37.2 & 22.4 & 12.6  & \checkmark \\
HotSpotNet \cite{hotspotnet} & 66.0 & 59.3 & 27.4 & 23.9 & 38.4 & 33.3 & 13.3 & \checkmark \\
CenterPoint \cite{centerpoint} & 65.5 & 58.0 & - & - & - & - & - & \checkmark \\
VISTA-OHS \cite{deng2022vista} & \textbf{69.8} & \textbf{63.0} & 25.6 & 23.3 & 32.1 & 21.6 & 12.2 & \checkmark \\
\midrule
Object-DGCNN (pillar) \cite{objectdgcnn} & 62.8 & 53.2 & 34.6 & 26.5 & 31.6 & 26.0 & 19.1 & \ding{55} \\ 
Object-DGCNN (voxel) \cite{objectdgcnn} & 66.0 & 58.7 & 33.3 & 26.3 & 28.8 & 25.1 & 19.0 & \ding{55} \\ 
Ours (pillar) & 63.0 & 53.8 & 35.1 & 26.4 & 32.1 & 26.5 & 19.0 & \ding{55} \\
Ours (voxel) & \textbf{67.6} & \textbf{61.3} & 30.5 & 25.4 & 35.2 & 26.7 & 12.5 & \ding{55} \\
\bottomrule
\end{tabular}
\end{table*}

We compare the recent works on KITTI \cite{kitti} dataset for \textit{car} category as shown in Table~\ref{tab:quantreskitti}. Our network achieves a competitive performance to the state-of-the-art LiDAR-based approaches like VoxelNet \cite{zhou2018voxelnet}, PointPillars \cite{pointpillars}, TANet \cite{liu2020tanet} and SECOND \cite{second} in terms of $AP_{3D}$ and $AP_{BEV}$ for \textit{easy}, \textit{moderate} and \textit{hard} samples. Our network could not achieve the state-of-the-art performance on KITTI \cite{kitti} dataset as compared to nuScenes \cite{caesar2020nuscenes} dataset because the transformer network is data hungry and  KITTI dataset has 3712 samples for training which is approximately 7.5 times less number of training samples than nuScenes dataset (which has 28K training samples). In addition to this, the nuScenes dataset provides 9 non-key frames (called sweeps) to aggregate to one key frame, resulting in dense LiDAR points but KITTI dataset provides only one LiDAR key frame which results in sparse point cloud. However, our approach obtains competitive performance to the state-of-the-art transformer based architecture VoTr-SSD \cite{votr} which uses NMS. To the best of our knowledge this is the first transformer based 3D detection network to report results both on nuScenes and KITTI datasets, which compares the results of detection with an emphasis on training sample size which is significant for transformer based architectures alongside the density of LiDAR point clouds. We further provide comparison of methods in terms of $AP_{3D}$ and $AP_{BEV}$ on the \textit{pedestrian} and \textit{cyclist} categories for \textit{easy}, \textit{moderate} and \textit{hard} samples in the supplementary.

\begin{table*}[htp]
    \centering
    \caption{Comparison of recent works in terms of $AP_{3D}$ and $AP_{BEV}$ detection on KITTI \cite{kitti} \textit{val} set. We list results for car category for \textit{easy}, \textit{moderate} and \textit{hard} samples with IoU=0.7.}
    \label{tab:quantreskitti}
    \begin{tabular}{c|ccc|ccc|c}
    \toprule
    & \multicolumn{3}{|c|}{$AP_{3D}$} & \multicolumn{3}{|c|}{$AP_{BEV}$} & \\
    Method & Easy & Mod. & Hard & Easy & Mod. & Hard & NMS \\ \midrule
    \multicolumn{8}{c}{\textit{RGB \& LiDAR}} \\ \midrule
    MV3D \cite{mv3d} & 71.2 & 62.6 & 56.5 & 86.5 & 78.1 & 76.6 & \checkmark \\
    AVOD-FPN \cite{avod} & - & 73.2 & - & - & - & - & \checkmark \\
    F-PointNet \cite{qi2018frustum} & 87.3 & 70.9 & 63.6 & 88.1 & 84.0 & 76.4 & \checkmark \\
    3D-CVF \cite{3dcvf} & 89.6 & 79.8 & 78.4 & - & - & - & \checkmark \\ \midrule
    \multicolumn{8}{c}{\textit{LiDAR}} \\ \midrule
    VoxelNet \cite{zhou2018voxelnet} & 81.9 & 65.4 & 62.8 & 88.0 & 78.4 & 71.3 & \checkmark \\
    PointPillars \cite{pointpillars} & 86.6 & 76.0 & 68.9 & 90.1 & 86.6 & 82.8 & \checkmark \\
    TANet \cite{liu2020tanet} & 87.5 & 76.6 & 73.8 & - & - & - & \checkmark \\
    SECOND \cite{second} & 87.4 & 76.4 & 69.1 & 89.4 & 83.8 & 78.6 & \checkmark \\
    3DSSD \cite{yang20203dssd} & 89.7 & 79.4 & 78.6 & \textbf{92.7} & \textbf{89.0} & \textbf{85.9} & \checkmark \\
    VoTr-SSD \cite{votr} & 87.8 & 78.2 & 76.9 & - & - & - & \checkmark \\
    Pointformer \cite{pointformer} & \textbf{90.0} & \textbf{79.6} & \textbf{78.8} & - & - & - & \checkmark \\
    Ours (voxel) & 87.6 & 76.8 & 73.9 & 89.6 & 86.8 & 83.1 & \ding{55}\\ \bottomrule
    \end{tabular}
\end{table*}

\subsubsection{Qualitative Results}
\begin{figure*}[htp]
    \centering
    \begin{subfigure}[b]{0.49\textwidth}
        \centering
        \includegraphics[width=\textwidth]{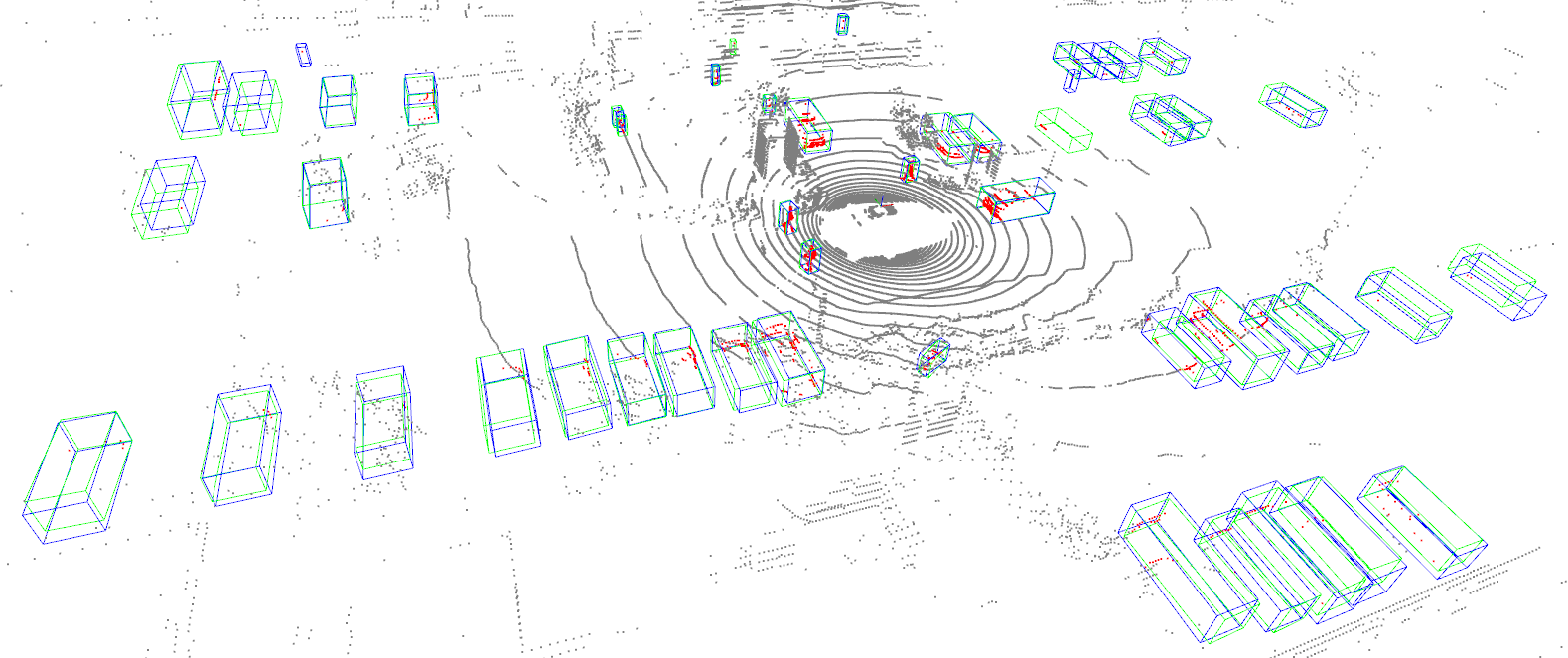}
    \end{subfigure}
    \begin{subfigure}[b]{0.49\textwidth}
        \centering
        \includegraphics[width=\textwidth]{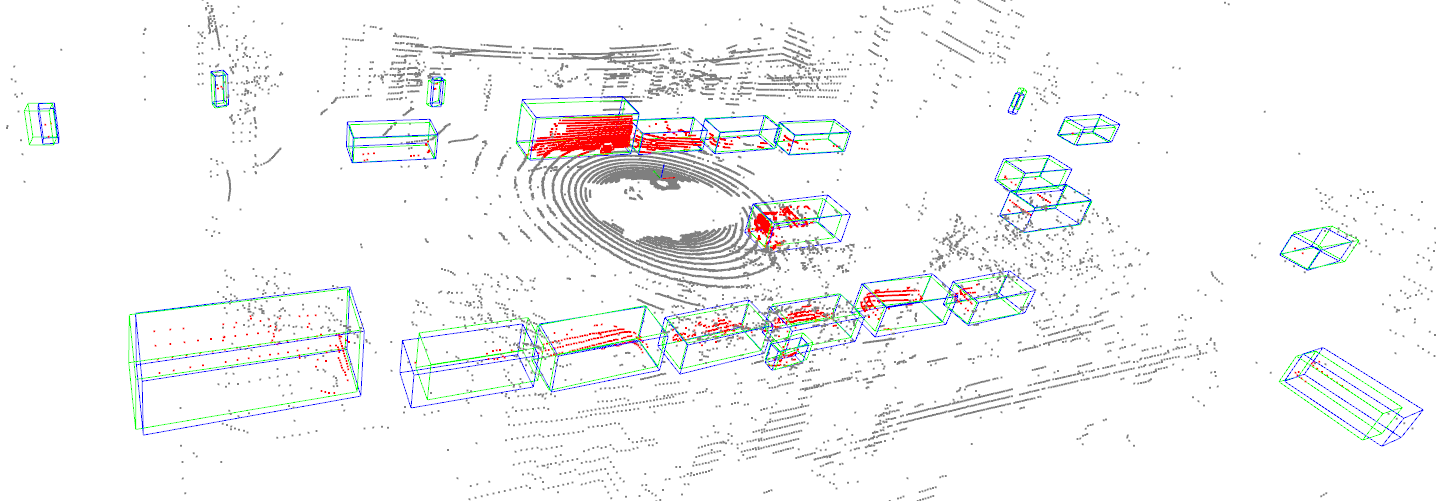}
    \end{subfigure} 
    \caption{Visualization of results on the nuScenes dataset. \textcolor{blue}{Blue} and \textcolor{green}{green} boxes represent predictions and ground-truth respectively. Points inside the bounding boxes are shown in \textcolor{red}{red}. Best viewed in color and zoom-in.}
    \label{fig:qualresults}
\end{figure*}

The visualization of 3D bounding box predictions of our approach on the nuScenes dataset is shown in Figure~\ref{fig:qualresults}. Although the LiDAR point clouds are sparse, our approach not only detects small objects like traffic cones but also efficiently detect large size objects like truck, bus, construction vehicle. This is possible with the local and global feature maps of backbone and attention mechanism in encoder alongside the cross-attention in the decoder. Our approach is also able to detect some cars which are not annotated in ground-truth. A short demo video of 3D object predictions of our network projected into BEV map is presented at \url{https://youtu.be/5pLnLRO_2-U}.

\subsubsection{Analysis}
\label{subsec:analysis}
The performance analysis of our approach by object category, object distance and object size compared to state-of-the-art LiDAR based CNN and transformer models is detailed below.

\noindent\textbf{Object category.}
The performance of our network in terms of Average Precision (AP) for each object category compared to other state-of-the-art networks on the nuScenes \cite{caesar2020nuscenes} \textit{val} dataset is shown in Table~\ref{tab:apbycat}. We compare with Object-DGCNN \cite{objectdgcnn} as it a \textit{standalone} transformer model like ours for fair comparison. The transformer encoder which extracts global LiDAR features leveraging long range interactions using multi-scale deformable attention and decoder cross-attention significantly improves the AP of large size objects like bus, construction vehicle, trailer and truck. Although the number of bicycle category objects are less compared to other objects, our model surpassed Object-DGCNN \cite{objectdgcnn} by 6.9\% AP, this is possible with the local and global feature extraction in addition to Li3DeTr cross-attention of decoder. The other object categories obtain a competitive performance. We quantize the point clouds in the backbone of our network and downsample the feature maps into multiple strides to increase the receptive field, but that leads to information loss which thereby makes our network difficult to detect smaller objects like \textit{pedestrians} and \textit{barriers}. The design of point cloud backbones in the future, maintaining the original resolution while increasing the receptive field would solve this problem. The performance of our network in terms of AP for each object category on nuScenes \textit{test} dataset is provided in supplementary.

\begin{table*}[htp]
\centering
\small{
\caption{Performance of our network in terms of Average Precision (AP) by object category on the nuScenes \textit{val} set. CV - Construction Vehicle, Motor - Motorcycle, Ped - Pedestrian, TC - Traffic Cone, Barr - Barrier. $\ast$: MMDetection3D \cite{mmdet3d2020} implementation. The scores in \textcolor{green}{green} indicate the increase in performance with respect to scores in \underline{underline}.}
\label{tab:apbycat}
\begin{tabular}{@{}l|llllllllll|l@{}}
\toprule
Method        & Car & Truck & Trailer & Bus & CV & Bicycle & Motor & Ped &TC &Barr & mAP \\ \midrule 
Pointformer \cite{pointformer} & 82.3 & 48.1 & 43.4 & 55.6 & 8.6 & 22.7 & 55.0 & 81.8 & 72.2 & 66.0 & 53.6\\
CenterPoint \cite{centerpoint} $\ast$ &85.1 &53.0 &35.4 &66.8 &13.9 &34.4 &55.2 &84.6 &66.9 &67.5 & 56.2 \\
VISTA \cite{deng2022vista} & 85.0 & 57.4 & 39.9 & 66.4 & 21.2 & 51.7 & 66.6 & 84.5 & 68.5 & 66.8 & 60.9 \\
Obj-DGCNN \cite{objectdgcnn} &84.0 &\underline{54.0} &\underline{40.4} &\underline{66.8} &\underline{20.2} &\underline{44.7} &66.2 &81.6 &64.7 &62.6 & 58.5\\
Ours &85.8 &56.5 \textcolor{green}{$\uparrow$ 2.5} &43.0 \textcolor{green}{$\uparrow$ 2.6} &70.9 \textcolor{green}{$\uparrow$ 4.1} &22.9 \textcolor{green}{$\uparrow$ 2.7} &51.6 \textcolor{green}{$\uparrow$ 6.9}  &66.9 &83.9 &66.8 &65.7 & \textbf{61.4}\\ \bottomrule
\end{tabular}
}
\end{table*}

\noindent\textbf{Object distance.}
The ground-truth 3D bounding boxes are divided into three subsets: \verb|[0m,20m]|, \verb|[20m,30m]| and \verb|[30m,+|$\infty$\verb|]| basing on the distance between object centers and ego vehicle. The performance of our network in terms of mAP by object distance on the nuScenes \cite{caesar2020nuscenes} dataset compared to  CenterPoint \cite{centerpoint} and Object-DGCNN \cite{objectdgcnn} is shown in Table~\ref{tab:mapbydistance}. Our approach significantly improves the mAP of objects at object distance greater than \verb|30m| compared to CNN-based CenterPoint \cite{centerpoint}. Although the LiDAR point cloud is sparse at far distance from ego vehicle, our attention mechanism in encoder and decoder models long range interactions between the sparse points to predict the objects at far distance.

\begin{table}[htp]
\centering
 \small{   
\caption{Performance of our network in terms of mAP by object distance on the nuScenes \textit{val} set. $\ast$: MMDetection3D \cite{mmdet3d2020} implementation. The scores in \textcolor{green}{green} indicate the increase in performance with respect to scores in \underline{underline}.}
\label{tab:mapbydistance}
\begin{tabular}{@{}l|lll@{}}
\toprule
Method        & \verb|[0m,20m]| & \verb|[20m,30m]| & \verb|[30m,+|$\infty$\verb|]| \\ \midrule 
CenterPoint \cite{centerpoint} $\ast$ &\underline{71.3} &51.5 &\underline{26.5}  \\
Obj-DGCNN \cite{objectdgcnn} &73.2 &55.5 &30.3 \\
Ours & 75.6 \textcolor{green}{$\uparrow$ 4.3} & 56.9  &32.7 \textcolor{green}{$\uparrow$ 6.2} \\ \bottomrule
\end{tabular}
}
\end{table}

\noindent\textbf{Object size.}
The ground-truth 3D bounding boxes are divided into two subsets: \verb|[0m, 4m]| and \verb|[4m,+|$\infty$\verb|]|, basing on the size of the longer edge of the bounding box. The performance of our model in terms of mAP by object size compared to state-of-the-art approaches on the nuScenes \cite{caesar2020nuscenes} dataset is shown in Table~\ref{tab:mapbysize}. Our transformer based approach predominantly improves the mAP of large size objects than small size objects compared to CNN-based CenterPoint \cite{centerpoint}. This proves our hypothesis that long range interactions made possible by attention mechanism improves the detection performance.

\begin{table}[htp]
\centering
\caption{Performance of our network in terms of mAP by object size on the nuScenes \textit{val} set. $\ast$: MMDetection3D \cite{mmdet3d2020} implementation. The scores in \textcolor{green}{green} indicate the increase in performance with respect to scores in \underline{underline}.}
\label{tab:mapbysize}
\begin{tabular}{@{}l|ll@{}}
\toprule
Method        & \verb|[0m,4m]| & \verb|[4m,+|$\infty$\verb|]| \\ \midrule 
CenterPoint \cite{centerpoint} $\ast$ &\underline{34.9} & \underline{23.5}  \\
Obj-DGCNN \cite{objectdgcnn} &36.0 &25.4   \\
Ours &37.9 \textcolor{green}{$\uparrow$ 3.0} & 27.8 \textcolor{green}{$\uparrow$ 4.3} \\ \bottomrule
\end{tabular}
\end{table}



\subsection{Ablation Studies}
\label{subsec:ablation}

\noindent \textbf{Attention blocks.} The performance of our network in terms of mAP and NDS with different attention operations for self and cross-attention blocks in the decoder (\S~\ref{subsec:decoder}) is shown in Table~\ref{tab:attn}. We test our approach by employing DGCNN \cite{dgcnn} similar to \cite{objectdgcnn} and multi-head self-attention \cite{vaswani2017attention} to model object query interactions and, deformable cross-attention \cite{zhu2020deformable} and our Li3DeTr cross-attention to attend the global voxel features. Our Li3DeTr cross-attention shows improved performance compared to deformable cross-attention \cite{zhu2020deformable} for both of the self-attention operations. This proves the effectiveness of our Li3DeTr cross-attention block to effectively link the global voxel features with the 3D object predictions.

\begin{table}[htp]
    \centering
        \small{
        \caption{Ablation study with different attention operations in the decoder on the nuScenes \textit{val} set}
        \label{tab:attn}
        \begin{tabular}{@{}cc|cc@{}}
        \toprule
        Self-attention & Cross-attention & mAP & NDS \\ \midrule
        \multirow{2}{*}{DGCNN \cite{dgcnn}} & Deformable attn.  \cite{zhu2020deformable} & 58.6 & 66.0 \\
        & Li3DeTr (ours) & \textbf{59.0} & \textbf{66.3} \\\midrule
        \multirow{2}{*}{Multi-head self attn. \cite{vaswani2017attention}} & Deformable attn. \cite{zhu2020deformable} & 57.9 & 65.5 \\
        &Li3DeTr (ours) & \textbf{61.4} & \textbf{67.6} \\ \bottomrule
        \end{tabular}
    }
\end{table}

\noindent \textbf{Number of queries.} The performance of our network in terms of mAP and NDS on the nuScenes \cite{caesar2020nuscenes} dataset for different number of queries in the decoder is shown in Table~\ref{tab:queryablation}. The performance of our network in terms of mAP and NDS slightly increases with increase in number of queries because object queries represent the potential positions of objects. However, the performance has slight impact for over 900 queries. So, we fix number of queries to 900.

\begin{table}[htp]
\centering
\caption{Ablation study on the number of object queries in decoder on the nuScenes \textit{val} set}
\label{tab:queryablation}
\begin{tabular}{@{}c|cc|cc@{}}
\toprule
 & \multicolumn{2}{|c|}{Li3DeTr (pillar)} & \multicolumn{2}{c}{Li3DeTr (voxel)} \\ \midrule
No. of queries & mAP & NDS &mAP &NDS \\ \midrule
300 &51.9 &60.4 &59.3 &65.1 \\
600 &52.5 &61.4 &60.2 &66.1 \\
900 &\textbf{53.8} &\textbf{63.0} &\textbf{61.4} &\textbf{67.6} \\
1000 &53.1 &62.5 &60.8 &67.2 \\\bottomrule
\end{tabular}
\end{table}

\noindent\textbf{Backbones.} The performance of our network in terms of mAP and NDS on nuScenes dataset for different backbones compared with CenterPoint \cite{centerpoint} and Object-DGCNN \cite{objectdgcnn} is shown in Table~\ref{tab:backboneablation}. We test our approach with PointPillars \cite{pointpillars} with 0.2m voxel size and VoxelNet \cite{zhou2018voxelnet} with 0.1m voxel size for LiDAR point cloud feature extraction. As shown in Table~\ref{tab:backboneablation}, our approach with VoxelNet feature extraction outperforms the network with PointPillars backbone. However, our architecture is flexible to plugin various backbones depending on the specific requirements of different applications.

\begin{table}[htp]
    \centering
            \caption{Ablation study on different backbones on the nuScenes \textit{val} set}
            \label{tab:backboneablation}
                \begin{tabular}{@{}cc|cc@{}}
                \toprule
                Backbone & Method & mAP & NDS \\ \midrule
                \multirow{3}{*}{PointPillars \cite{pointpillars}} & CenterPoint \cite{centerpoint} & 50.3 & 60.2 \\
                &Object-DGCNN \cite{objectdgcnn} & 53.2 & 62.8 \\
                & Ours & \textbf{53.8} & \textbf{63.0} \\ \midrule
                \multirow{3}{*}{VoxelNet \cite{zhou2018voxelnet}} & CenterPoint \cite{centerpoint} & 56.4 & 64.8 \\
                &Object-DGCNN \cite{objectdgcnn} & 58.6 & 66.0 \\
                & Ours & \textbf{61.4} & \textbf{67.6} \\ \bottomrule
                \end{tabular}
\end{table}

\section{Conclusion}
We present an end-to-end, single-stage LiDAR based 3D Detection Transformer (Li3DeTr) architecture which inputs LiDAR point clouds and predicts 3D bounding boxes. Inspired by DETR \cite{detr}, we formulate our model with set-to-set loss and thereby remove the need for post processing methods like NMS. We introduce a novel Li3DeTr cross-attention block in the decoder head to link the global LiDAR voxel feature maps (obtained from encoder network) and 3D predictions, leveraged by sparse set of object queries learnt from the data. Without bells and whistles, our network archives 61.3\% mAP and 67.6\% NDS surpassing the state-of-the-art methods on the nuScenes dataset and achieves competitive performance on the KITTI dataset. 

\section*{Acknowledgment}
This work has been supported by European Union's H2020 MSCA-ITN-ACHIEVE with grant agreement No. 765866, Fundação para a Ciência e a Tecnologia (FCT) under the project UIDB/00048/2020 and FCT Portugal PhD research grant with reference 2021.06219.BD.

{\small
\bibliographystyle{ieee_fullname}
\bibliography{egbib}
}

\end{document}


\title{Li3DeTr: A LiDAR based 3D Detection Transformer\\
Supplementary Material}

\author{Gopi Krishna Erabati and Helder Araujo\\
Institute of Systems and Robotics\\
University of Coimbra, Portugal\\
{\tt\small \{gopi.erabati, helder\}@isr.uc.pt}
}

\maketitle
\thispagestyle{empty}

\section{Implementation Details}
\subsection{Model Details}
Our model mainly consists of two modules: CNN backbone and transformer encoder-decoder. We test with SparseConv \cite{sparseconv} and PointPillars \cite{pointpillars} with SECOND \cite{second} as feature extraction network. nuScenes dataset: The point cloud range is set to $[-51.2m, 51.2m] \times [-51.2m, 51.2m] \times [-5.0m, 3.0m]$. We employ SparseConv with voxelization resolution set to $[0.1m, 0.1m, 0.2m]$ and four blocks of [3, 3, 3, 2] 3D sparse convolutions with [16, 32, 64, 128] dimensions and PointPillars with voxelization resolution set to $[0.2m, 0.2m, 8m]$ and three blocks of [3, 5, 5] convolutional layers with [64, 128, 256] dimensions. KITTI dataset: The point cloud range is set to $[0.0m, 71.4m] \times [-40.0m, 40.0m] \times [-3.0m, 1.0m]$. We employ SparseConv with voxelization resolution set to $[0.05m, 0.05m, 0.1m]$ and four blocks of [1, 3, 3, 3] 3D sparse convolutions with [16, 32, 64, 64] dimensions and PointPillars with voxelization resolution set to $[0.16m, 0.16m, 4.0m]$ and three blocks of [3, 5, 5] convolutional layers with [64, 128, 256] dimensions. We use FPN \cite{fpn} to transform the features and obtain four multi-scale local LiDAR feature maps with 256 channel dimension. For the transformer encoder, we use $L_{enc}=2$ encoder layers with multi-scale deformable self-attention \cite{zhu2020deformable} with [256, 256] dimensions, 8 heads, 4 levels and 4 sampling points for each query and in each head. For the decoder, we use $L_{dec}=6$ decoder layers with hidden dimension 256 and 900 object queries.

\subsection{Training Details}
We implement our model in PyTorch \cite{pytorch} based on open-sourced MMDetection3D \cite{mmdet3d2020}. We train Li3DeTr network with AdamW optimizer with initial learning rate of $2 \times 10^{-4}$ and weight decay of $10^{-2}$. Cosine Annealing is set as learning rate scheduler with 2K warm up iterations and with minimum learning rate of $2 \times 10^{-6}$. The backbone is initialized with pretrained network \cite{objectdgcnn, second} on the same dataset. The model is trained for 30 epochs on two RTX 3090 GPUs with a batch size of 4. During test time, we take the top 300 predictions with highest category score as final predictions and we do not use any NMS.

\section{More Quantitative Results}
The performance of our Li3DeTr network compared with other state-of-the-art approaches on the nuScenes \textit{val} dataset in terms of mAP and NDS is shown in Table~\ref{tab:quantnusval}. Our network outperforms all the methods in terms of mAP and stands second in terms of NDS on nuScenes \textit{val} set. Although VISTA \cite{deng2022vista} uses NMS to remove redundant boxes, our method achieves improved performance in 3D object detection without NMS. Moreover VISTA is a plug and play module, but our approach is a \textit{standalone} network for 3D object detection. Our network surpassed the state-of-the-art transformer based NMS-free network Object-DGCNN \cite{objectdgcnn} by 2.8 \% mAP and 1.6 \% NDS.

\begin{table}[htp]
    \centering
        \small{
        \caption{Comparison of recent works in terms of mAP and NDS on the nuScenes \cite{caesar2020nuscenes} \textit{val} set. The scores in \underline{underline} represent \textit{second} position in the corresponding metrics.}
        \label{tab:quantnusval}
        \begin{tabular}{@{}c|ccc@{}}
        \toprule
        Method  & NDS $\uparrow$ & mAP $\uparrow$ & NMS \\ \midrule
        CenterPoint \cite{centerpoint} & 64.8 & 56.4 & \checkmark \\
        HotSpotNet \cite{hotspotnet} & 66.0 & 59.5 & \checkmark \\
        VISTA-OHS \cite{deng2022vista} & \textbf{68.1} & \underline{60.8} & \checkmark \\
        Object-DGCNN (voxel) \cite{objectdgcnn} &66.0  &58.6 & \ding{55} \\
        Ours (voxel) & \underline{67.6}  &\textbf{61.4} & \ding{55} \\\bottomrule
        \end{tabular}
    }
\end{table}

The performance of our network compared with state-of-the-art approaches on the KITTI \cite{kitti} \textit{val} dataset for \textit{pedestrian} and \textit{cyclist} categories is shown in Table~\ref{tab:quantreskittipedcyc}. The results of our approach on the \textit{car} category shows competitive performance with state-of-the-art approaches as shown in Table~2 (in main paper). However, our method could not achieve state-of-the-art performance on \textit{pedestrian} and \textit{cyclist} categories. The number of object samples in the training split of KITTI \cite{kitti} dataset for \textit{car} category is 83\%, whereas for \textit{pedestrian} and \textit{cyclist} categories is 13\% and 4\% respectively. Due to very less number of object samples during training, our transformer network could not achieve competitive results on \textit{pedestrian} and \textit{cyclist} categories. In addition to this, as described in \S~4.2.3 (in main paper), due to quantization of point clouds and downsampling of feature maps, our network finds it difficult to detect small size objects.

\begin{table*}[htp]
    \centering
    \small{ 
    \caption{Comparison of recent works in terms of $AP_{3D}$ and $AP_{BEV}$ detection on KITTI \cite{kitti} \textit{val} set. We list results for \textit{pedestrian} and \textit{cyclist} category for \textit{easy}, \textit{moderate} and \textit{hard} samples with IoU=0.5. The scores in \underline{underline} represent second rank in the corresponding metric.}
    \label{tab:quantreskittipedcyc}
    \begin{tabular}{c||ccc|ccc||ccc|ccc||c}
    \toprule
    & \multicolumn{6}{c||}{Cyclist} & \multicolumn{6}{c||}{Pedestrian} & \\ \cline{2-7} \cline{8-13}
    & \multicolumn{3}{c|}{$AP_{3D}$} & \multicolumn{3}{c||}{$AP_{BEV}$} & \multicolumn{3}{c|}{$AP_{3D}$} & \multicolumn{3}{c||}{$AP_{BEV}$} & \\ \cline{2-4} \cline{5-7} \cline{8-10} \cline{11-13}
    Method & Easy & Mod. & Hard & Easy & Mod. & Hard & Easy & Mod. & Hard & Easy & Mod. & Hard & NMS \\ \midrule
    F-PointNet \cite{qi2018frustum} & 77.1 & 56.4 & 53.3 & \textbf{81.8} & 60.0 & 56.3 & \underline{70.0} & \underline{61.3} & \underline{53.5} & \textbf{72.3} & \textbf{66.3} & \textbf{59.5} & \checkmark \\
    VoxelNet \cite{zhou2018voxelnet}  & 67.1 &47.6 & 45.1 & 74.4 &52.1 & 50.4 & 57.8 & 53.4 & 48.8 & \underline{65.9} & \underline{61.0} & \underline{56.9} & \checkmark  \\
    PVCNN \cite{pvcnn}  & \textbf{81.4} & 59.9 & 56.2 & - & - & - & \textbf{73.2} & \textbf{64.7} & \textbf{56.7} & - & - & - &  \checkmark \\
    Complex-YOLO \cite{simony2018complexyolo}  & 68.1 & 58.3 & 54.3 & 72.3 & \underline{63.3} & \underline{60.2} & 41.7 & 39.7 & 35.9 & 46.0 & 45.9 & 44.2 & \checkmark \\
    PointPillars \cite{pointpillars}  & 80.0 & \underline{62.6} & \underline{59.5} & - & - & - & 57.7 & 52.2 & 47.9 & - & - & - & \checkmark \\
    SECOND \cite{second}  & \underline{80.5} & \textbf{67.1} & \textbf{63.1} &  - & - & - & 56.5 & 52.9 & 47.7 & - & - & - & \checkmark \\
    Ours  & 77.0 &60.0 & 58.0 & \underline{80.5} &\textbf{64.0}	&\textbf{60.3} & 51.1	& 44.1	&40.0	& 56.5	& 50.0 & 45.2 &\ding{55} \\
    \bottomrule
    \end{tabular}
    }
\end{table*}

\section{More Analysis and Ablation Studies}

\subsection{More Analysis on Object Category}
The performance of our network in terms of Average Precision (AP) for each object category compared to other state-of-the-art networks on the nuScenes  \cite{caesar2020nuscenes} \textit{test} dataset is shown in Table~\ref{tab:apbycattest}. Similar to results on nuScenes \textit{val} dataset, the global voxel features by multi-scale deformable attention \cite{zhu2020deformable} and our novel cross-attention block significantly improves the AP of large size objects like trailer, bus, construction vehicle compared with Object-DGCNN \cite{objectdgcnn}. However, the performance on smaller objects like pedestrian and barrier is worse due to quantization of point cloud and downsampling of feature maps in the backbone to increase the receptive filed which results in information loss.

\begin{table*}[htp]
\centering
\small{
\caption{Performance of our network in terms of Average Precision (AP) by object category on the nuScenes \textit{test} set. CV - Construction Vehicle, Motor - Motorcycle, Ped - Pedestrian, TC - Traffic Cone. $\ast$: MMDetection3D \cite{mmdet3d2020} implementation. The scores in \textcolor{green}{green} indicate the increase in performance with respect to scores in \underline{underline}.}
\label{tab:apbycattest}
\begin{tabular}{@{}l|llllllllll|l@{}}
\toprule
Method        & Car & Truck & Trailer & Bus & CV & Bicycle & Motor & Ped &TC &Barrier & mAP \\ \midrule 
CenterPoint \cite{centerpoint} $\ast$ &84.6 &51.0 &53.2 &60.2 &17.5 &28.7 &53.7 &83.4 &76.7 &70.9 & 58.0 \\
VISTA \cite{deng2022vista} & 84.4 & 55.1 & 54.2 & 63.7 & 25.1 & 45.4 & 70.0 & 82.8 & 78.5 & 71.4 & \textbf{63.0} \\
    Obj-DGCNN \cite{objectdgcnn} &\underline{84.0} &48.5 &\underline{54.0} &\underline{57.5} &\underline{25.2} &\underline{32.2} &64.5 &81.7 &73.8 &65.6 & 58.7 \\
Ours &85.6 \textcolor{green}{$\uparrow$ 1.6} &50.0 &56.5 \textcolor{green}{$\uparrow$ 2.5} &60.3 \textcolor{green}{$\uparrow$ 2.8} &30.3 \textcolor{green}{$\uparrow$ 5.1} &38.3 \textcolor{green}{$\uparrow$ 6.1}  &65.9 &83.0 &75.5 &68.0 & 61.3\\ \bottomrule
\end{tabular}
}
\end{table*}

\subsection{More Analysis on Inference Time}
The inference speed of our approach is compared with other state-of-the-art methods as shown in Table~\ref{tab:inferencefps}. Our approach not only achieves improvement in performance but also is faster than CenterPoint \cite{centerpoint} and Object-DGCNN \cite{objectdgcnn}.

\begin{table}[htp]
    \centering
        \caption{Comparison of inference speed measured on a NVIDIA RTX 3090 GPU. $\ast$: MMDetection3D \cite{mmdet3d2020} implementation}
        \label{tab:inferencefps}
    \begin{tabular}{@{}c|c|c|c@{}}
        \toprule
        Method & CenterPoint $\ast$ & Object-DGCNN & Ours \\ \midrule
        FPS & 5.1 & 5.6 & \textbf{7.1} \\ \bottomrule
    \end{tabular}
\end{table}

\subsection{Ablation on Detection Layers} The performance of our network in terms of mAP, NDS and other TP metrics on the nuScenes \cite{caesar2020nuscenes} dataset for different number of layers in the transformer decoder is shown in Table~\ref{tab:detlayers}. The hypothesis that we iteratively refine the object queries after each decoder layer to significantly improve the performance of the model is proved to be correct as shown in Table~\ref{tab:detlayers}. The performance significantly improves as we increase the number of decoder layers in the network. However, the performance of the model does not improve much after 6 decoder layers, so we fix number of decoder layers ($L_{dec}$) to 6.

\begin{table}[htp]
\centering
    \small{
\caption{Performance of our network on the nuScenes \textit{val} set by number of decoder layers. Among TP metrics only mATE, mASE, mAOE are shown.}
\label{tab:detlayers}
\begin{tabular}{c|ccccccc}
\toprule

Layer &NDS $\uparrow$ & mAP $\uparrow$ & mATE $\downarrow$ & mASE $\downarrow$ & mAOE $\downarrow$ \\ \midrule
1   &63.1   &54.5   &37.9   &27.1   &30.1   \\ 
2	&65.7   &58.4   &34.2   &26.6   &29.5   \\
3	&67.0   &60.6   &32.9   &26.6   &28.7   \\
4	&67.4   &61.2   &32.8   &26.5   &28.6   \\
5	&67.5   &61.3   &32.7   &26.2   &28.5   \\
\textbf{6}	&\textbf{67.6}   &\textbf{61.4}   &\textbf{32.7}   &\textbf{26.1}   &\textbf{28.5}   \\ 
\bottomrule
\end{tabular}
}
\end{table}

{\small
\bibliographystyle{ieee_fullname}
\bibliography{egbibsupl}
}